\documentclass[
]{ceurart}

\usepackage{graphicx}
\usepackage{arydshln}
\usepackage{multirow}
\usepackage{listings}
\begin{document}

\copyrightyear{2022}
\copyrightclause{Copyright for this paper by its authors.
  Use permitted under Creative Commons License Attribution 4.0
  International (CC BY 4.0).}

\conference{ReNeuIR'23: Workshop on Reaching Efficiency in Neural Information Retrieval}

\title{Attention over pre-trained Sentence Embeddings for Long Document Classification}


\author[1,2]{Amine Abdaoui}[%
orcid=0000-0002-6160-8461,
email=amin.abdaoui@oracle.com
]
\cormark[1]
\address[1]{Huawei Ireland Research Center, Dublin, Ireland}
\address[2]{Oracle, Paris, France}

\author[1]{Sourav Dutta}[%
orcid=0000-0002-8934-9166,
email=surav.dutta2@huawei.com
]

\cortext[1]{This co-author is currently employed by Oracle but this work was conducted when he was at the Huawei Ireland Research Center.}

\begin{abstract}
Despite being the current de-facto models in most NLP tasks, transformers are often limited to short sequences due to their quadratic attention complexity on the number of tokens. 
Several attempts to address this issue were studied, either by reducing the cost of the self-attention computation or by modeling smaller sequences and combining them through a recurrence mechanism or using a new transformer model. 
In this paper, we suggest to take advantage of pre-trained sentence transformers to start from semantically meaningful embeddings of the individual sentences, and then combine them through a small attention layer that scales linearly with the document length. 
We report the results obtained by this simple architecture on three standard document classification datasets. 
When compared with the current state-of-the-art models using standard fine-tuning, the studied method obtains competitive results (even if there is no clear best model in this configuration). 
We also showcase that the studied architecture obtains better results when freezing the underlying transformers. A configuration that is useful when we need to avoid complete fine-tuning (e.g. when the same frozen transformer is shared by different applications). Finally, two additional experiments are provided to further evaluate the relevancy of the studied architecture over simpler baselines. 
\end{abstract}


\maketitle

\section{Introduction}

The Transformer model \cite{NIPS2017_Vaswani} is now established as the standard architecture in Natural Language Processing (NLP). Several variants of the original model achieved state-of-the-art results in many tasks \cite{bert, roberta, xlnet} including document classification. 
In addition to their accurate results, transformers are also efficient when compared to recurrent neural network encoders. However, this efficiency drops significantly on long sequences. Indeed, transformers compute \(n*n\) self-attention matrices to get the contextualized representations. Therefore, the memory and computational requirements grow-up quadratically with the number of tokens \(n\). For this reason, most transformer-based models are limited to a fixed number of tokens (usually 512 tokens). 

To address this limitation, several attempts were made to improve the transformer efficiency on longer sequences. A first family of methods tries to simplify the self-attention complexity by reducing the number of computed weights.  Concretely, instead of letting each token attend to every other token in the sequence, these methods restrict the computation of the attention weights to a small number of locations \cite{longformer, Routing}. Another popular approach consists in splitting the long input into smaller chunks that can be modeled more efficiently with a transformer. Then, the obtained representations can be combined using a recurrent neural network or another document-level transformer \cite{RoBERT, smith}. Finally, instead of combining the outputs of different chunks using a new model, \cite{transformerxl} and \cite{xlnet} implemented a caching mechanism that allows the first tokens of chunk \(i\) to have access to the hidden states of the last tokens of chunk \(i-1\).


In this paper, we suggest to take advantage of {\em pre-trained sentence transformers} to get meaningful sentence representations without the need of any further pre-training \cite{sentence-bert}. The availability and variety of these models allow to easily adapt our framework to different domains and languages\footnote{\url{https://github.com/UKPLab/sentence-transformers}}. 
Based on these sentence representations, we evaluate the use of a small {\em attention layer} to form a document representation by giving higher weights to more important sentences. 
Note that we do not compute full self-attention matrices between all sentence pairs but only attention weights between the unique document representation and the different sentence embeddings. Indeed, we believe that sentence representations are less sensitive to external context than token embeddings. Similar architectures that also use linear weighted aggregations were evaluated on other tasks \cite{li2020parade, althammer2022parm}. 

To evaluate these assumptions in the case of long document classification, we compare our proposed architecture with the current state-of-the-art models on three standard datasets. To our knowledge, this is the first detailed evaluation of these models on the same datasets. In addition to complete fine-tuning, we include a setting where the underlying transformers are frozen. Such scenario might be useful when the same transformer is shared by different applications (each application trains only its own task-specific layers). 

\section{Related Work}

Most of the work that tried to adapt transformers to long documents were evaluated on language generation \cite{reformer, transformerxl, Routing}. In this section, we will mainly focus on methods than can be used in language understanding tasks such as classification. 

The easiest way to deal with long sequences is to truncate them at the maximum sequence length supported by the model. Usually, the first 512 tokens are used and the following ones are just thrown away. Therefore, the first baseline in this paper will be simple truncation using the Roberta model \cite{roberta}, which is a widely used transformer for Natural Language Understanding. Furthermore, Roberta was used to initialize two other models that are also included in our experiments.

A more sophisticated approach uses sparse self-attention matrices to reduce the transformer complexity. Instead of computing all the \(n*n\) weights in each matrix, the idea is to compute only the ones that convey important relationships. For example, Longformer \cite{longformer} combines a windowed local attention for all tokens with a global attention for few important tokens. On the one hand, the authors proposed to compute attention weights between each token and all its neighbors that are included in a fixed window. On the other hand, they allow important tokens (e.g. [CLS]) to attend to the whole sequence of \(n\) tokens. Thanks to these optimisations, the authors were able to pre-train the Longformer model, which is able to handle 4096 tokens, starting from Roberta weights. Another work suggested to choose the computed attention weights dynamically based on the content \cite{Routing}. However, the model has been designed for character-level language generation as most of the other sparse attention methods. Therefore, we will only include Longformer in our evaluations to represent this family of methods. 

Another popular research direction is to use a hierarchical architecture in order to reduce the cost of the self-attention computation. Instead of applying one transformer to the whole sequence, the idea is to stack multiple models that handle a smaller number of inputs. Since transformer's complexity is \(O(n^2)\), applying multiple transformers to smaller sequences is better than applying one transformer to the whole sequence. Several studies that used multiple levels of transformers or a recurrent neural network on top of transformers have been proposed \cite{RoBERT, wu2021hi}. However, most of them have not been shared publicly with the community. 
SMITH, which is able to handle 2048 tokens, is the one of the rare pre-trained hierarchical models that is available online \cite{smith}. 
The proposed architecture is composed of two levels of abstraction: a sentence-level and a document-level. Each level uses a small transformer that has 4 and 3 layers respectively, 4 attention heads and 256 hidden dimensions. Therefore, the resulting model has less parameters than all the other models studied here which follow the common base architectures (12 layers, 12 attention-heads and 768 hidden dimensions). It was pre-trained using the usual masked word prediction and a novel masked sentence prediction task. Then, it was fine-tuned for document matching using a siamese architecture. In this paper, SMITH will be considered as a baseline in our experiments despite of its small size. To our knowledge, this is the first evaluation of SMITH on the document classification datasets considered. 

Finally, \cite{transformerxl} proposed TransformerXL which is able to model an unlimited number of tokens. The proposed auto-regressive model is also applied to smaller chunks extracted from the original long documents. However, the modeling is not conducted independently on each chunk. At each time step, the previous hidden states are reused to compute the current ones introducing a sort of memory that propagates across the different segments. Moreover, the usual absolute position embeddings were replaced by relative positional encoding in order to avoid confusion on token positions when handling different segments. The same authors also pre-trained XLNet \cite{transformerxl} in order to improve auto-regressive models in NLU tasks. In addition to handling an unlimited sequence length (thanks to the caching mechanism and relative positional encoding proposed in TransformerXL), the authors used permutation language modeling to capture bidirectional context when pre-training the model. For all these reasons, we will include XLNet as a baseline in our experiments.  

In this paper, we will compare the above mentioned baselines with an attention-based architecture that relies on pre-trained sentence transformers. These models are usually trained using a siamese architecture on sentence pair datasets to derive semantically meaningful sentence representations \cite{sentence-bert}. To our knowledge, this is the first attempt to use pre-trained sentence transformers for handling long  documents.

\section{Methods}

In this section, we will detail the studied {\em Attention over Sentence Embeddings} (AoSE) architecture and compare its complexity and size with existing baselines. 

\subsection{AoSE Architecture}

First, long documents are segmented into sentences using common sentence separators (full stop, line break, etc.). We define a minimum and a maximum number of tokens to avoid generating very small and very long segments that do not correspond to real sentences. Then, a sentence transformer will be used to map each sentence to a fixed dense representation $s_i$. Relying on such pre-trained models that are already geared towards producing meaningful sentence embeddings is certainly an important advantage. 
After that, we use an attention layer to combine the normalized sentence embeddings $s_i$ while giving higher weights to important sentences \cite{yang2016hierarchical}. To calculate these 
weights $\alpha_i$, we rely on a small neural network $W_s$ and a trainable context vector $u_s$ that is equivalent to the query in the Transformer's self-attention definition \cite{NIPS2017_Vaswani}. 

\begin{equation}
u_i =  tanh(W_s \times s_i + b_s)
\end{equation}

\begin{equation}
\alpha_i =  \frac{\exp(u_i^T \times u_s)}{\sum_{j=1}^{t} \exp(u_j^T \times u_s)}
\end{equation}

The document representation $v$ is then computed using a weighted sum of the different sentence embeddings $s_i$. 

\begin{equation}
v =  \sum_{i=1}^{t} \alpha_i  \times s_i
\end{equation}

Finally, a dense layer can be added on top of the document embedding $v$ in order to perform classification.
All these steps are presented in Figure \ref{fig:archi} below.

\begin{figure*}
  \includegraphics[width=\textwidth]{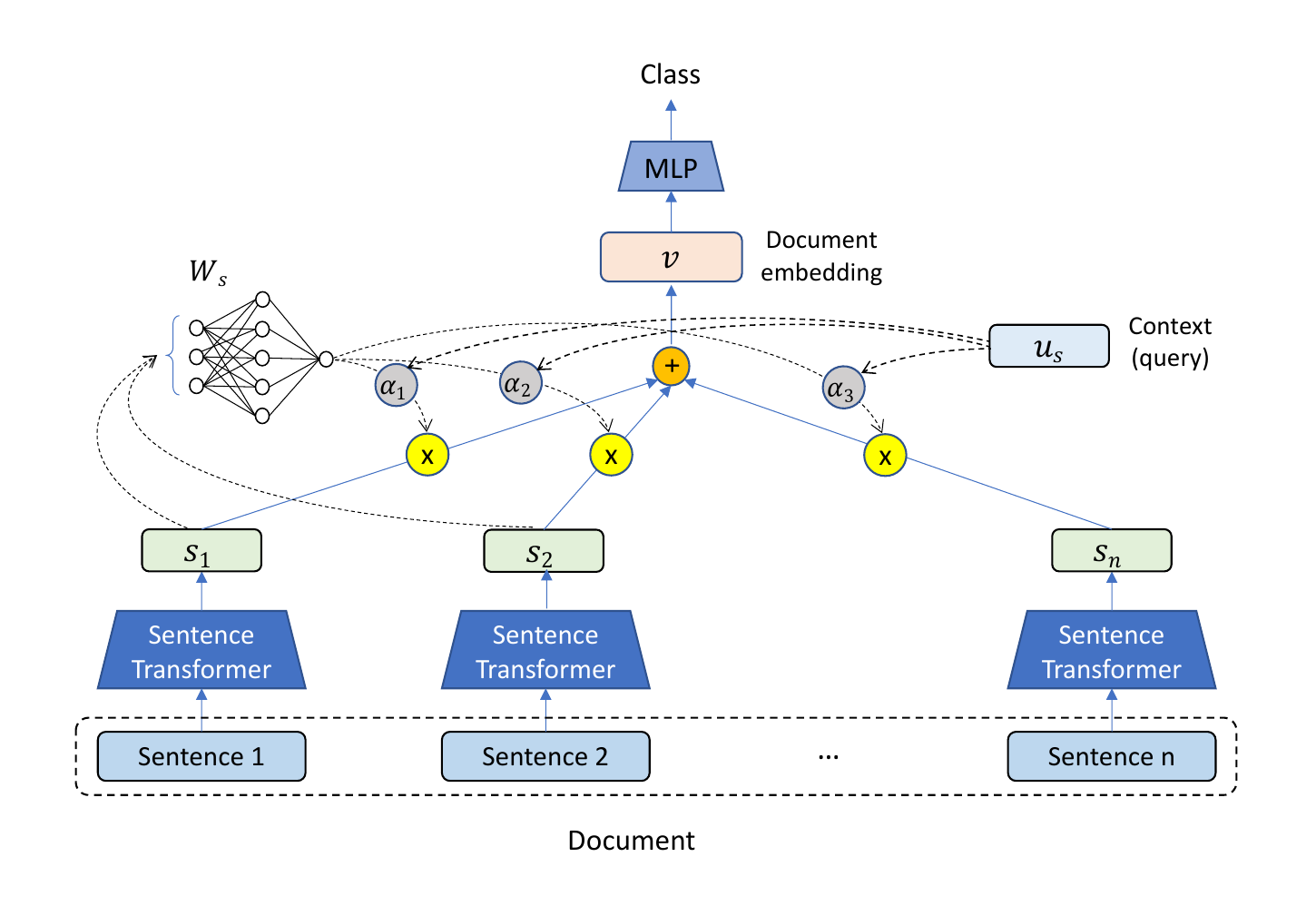}
  \caption{The proposed Attention over Sentence Embeddings (AoSE) architecture.}
  \label{fig:archi}
\end{figure*}

\subsection{Model Complexity}

Let's define the sequence length \(n\) as the product of the number of sentences \(t\) and the length of one sentence \(l\). The complexity of a vanilla transformer (e.g. Roberta) is therefore \(O(t^2 \times l^2)\). 

The sentence transformer of SMITH computes full self-attention between all tokens in each sentence, while the document transformer applies a second full self-attention computation between all sentences. Thus, the complexity of the whole SMITH encoder is \(O(t \times l^2 + t^2)\). 

Longformer computes global attention for \(g\) important tokens and local attention for the remaining ones.  Let \(w\) be the window size for local attention tokens. The Longformer's complexity is therefore \(O(g \times t \times l + (t \times l - g) \times w)\). 

XLNet computes full self-attention for each chunk. Let \(c\) be the maximum length of one XLNet segment (chunk). Even if all the previous hidden states are cached and reused, a given token can't attend to more than \(c\) locations. Therefore, the complexity of one chunk is \(O(c^2)\) and the number of chunks is \(t \times l / c\). Consequently, the complexity of XLNet is \(O(t \times l \times c)\).

The sentence transformer of our proposed architecture is equivalent to the one used in SMITH, but since our document-level attention is linear with the number of sentences, the complexity of our architecture is \(O(t \times l^2 + t)\).


\begin{table}
\caption{Complexity of the different models (\(t\) is the number of sentences, \(l\) is the average number of tokens per sentence, \(g\) is the number of global attention tokens, \(w\) is the window size for local attention tokens, and c is the length of one XLNet segment).}
\begin{tabular}{p{0.2\textwidth}c}
\hline
\textbf{Model} & \textbf{Complexity} \\
\hline
Roberta & \(O(t^2 \times l^2)\) \\
SMITH & \(O(t \times l^2 + t^2)\) \\
Longformer & \( O(g \times t \times l + (t \times l-g) \times w) \) \\
XLNet & \(O(t \times l \times c)\) \\
AoSE & \(O(t \times l^2 + t)\) \\ \hline
\end{tabular}
\label{tab:complexity}
\end{table}

\subsection{Model Size}

Due to hardware limitations, we decided to work with the base versions of Roberta, Longformer and XLNet even if large versions were also available. Similarly, we have chosen a base sentence transformer\footnote{\url{https://huggingface.co/sentence-transformers/nli-roberta-base-v2}} in our AoSE architecture. The chosen sentence transformer was initialized with Roberta base. Therefore, these two models share the exact same number of parameters. SMITH is the only exception as its only available version is much smaller in size. Table \ref{tab:size} presents several size-related measurements for all the models evaluated in this work.


We can also notice that our AoSE architecture has slightly more parameters than Roberta. As mentioned before, our architecture is composed of a sentence transformer (that has the same size as Roberta) and a small attention layer that has less than 1 million parameters. Most of them are located in the $W_s$ matrix that has a shape of $768 \times 768$. Therefore, our proposed architecture does not add a lot of parameters when compared to a standard transformer.

\begin{table*}
\caption{Different size comparisons of the evaluated models.}
\begin{tabular}{p{0.16\textwidth}ccccccc}
\hline
\textbf{Model} & \textbf{Disk} & \textbf{ \#Params } & \textbf{ Vocab.} & \textbf{ \#Layers / } & \textbf{ Hidden } & \textbf{ Max input} \\
  & \textbf{size} & (million) & \textbf{size} & \textbf{\#Heads} & \textbf{dim.} & \textbf{\#tokens} \\
\hline
Roberta & 478 MB & 125 m & 50.265 & 12 / 12 & 768 & 512 \\
SMITH & 47 MB & 12 m & 30.522 & 4+3 / 4 & 256 & 2.048 \\
Longformer & 570 MB & 149 m & 50.265 & 12 / 12 & 768 & 4.096 \\
XLNet & 445 MB & 117 m & 32.000 & 12 / 12 & 768 & unlimited \\
AoSE & 480 MB & 126 m & 50.265 & 12 / 12 & 768 & unlimited \\ \hline
\end{tabular}
\label{tab:size}
\end{table*}

\section{Experiments}

In this section, we will assess the performance of our architecture along with the selected baselines for long document classification. 

\subsection{Datasets}

Three classification datasets (of several thousands of documents each) were chosen to conduct our experiments. The first one is the widely used IMDB dataset \cite{maas2011learning}. We used the binary version\footnote{\url{https://huggingface.co/datasets/imdb}} that distinguishes positive and negative movie reviews. The second one is MIND \cite{wu2020mind}, a large-scale dataset for news recommendation. We used the topic classification task from this dataset\footnote{\url{https://msnews.github.io}} and discarded a couple of topics that have less than 3 documents. The third one is the 20 News Groups dataset \cite{newsweeder}.  We used the cleaned version\footnote{\url{https://huggingface.co/datasets/SetFit/20\_newsgroups}} that do not contain headers, signatures, and quotations. Table \ref{tab:stats} below shows several statistics related to the number of documents, the length of these documents and the number of classes for each dataset. The only additional preprocessing step that was applied to these datasets consisted in removing HTML tags. 

\begin{table*}
\caption{Statistics of the different datasets: (i) the total number of documents, (ii) the number of long documents (having more than 512 tokens), (iii) the average number of tokens per document, (iv) the maximum number of tokens per document, and (v) the number of labels. The number of tokens were computed using the roberta tokenizer.}
\begin{tabular}{lcccccc}
\hline
\textbf{Dataset} & \textbf{ Split } & \textbf{ \#Docs } & \textbf{ \#Docs } & \textbf{Avg} & \textbf{Max} & \textbf{\#Labels} \\
 & & \textbf{ all} & \textbf{ $>$512 } & \textbf{ \#Tokens } & \textbf{ \#Tokens } & \\
\hline
\multirow{2}{*}{IMDB} & train & 25.000 & 7.729 & 323 & 3.240 & 2  \\
 & test & 25.000 & 3.537 & 314 & 3.257 & 2 \\ \hline
\multirow{2}{*}{MIND} & train & 101.523 & 48.093 & 696 & 51.662 & 15  \\
 & test & 28.275 & 13.160 & 651 & 28.287 & 15 \\ \hline
20 News & train & 11.314 & 1.245 & 398 & 49.561 & 20  \\
 Groups & test & 7.532 & 784 & 372 & 132.115 & 20 \\ \hline
\end{tabular}
\label{tab:stats}
\end{table*}

\subsection{Experimental Settings}
\label{sec:settings}

Even if XLNet and AoSE can theoretically handle sequences of unlimited length\footnote{There is no structural limitation caused by the model definition (for example, the size of the position embeddings matrix).}, we had to set a maximum sequence length to each one due to practical hardware restrictions. Indeed, we were not able to fine-tune XLNet on very long sequences using our $2 \times 16$ GB GPUs (even with a batch size of 1). Therefore, we decided to set the maximum sequence length of XLNet to 4096 tokens in our experiments (which covers all IMDB and more than 99\% of MIND and 20 News Groups). In the case our AoSE model, we were able to input up to 8192 tokens which covers almost all the documents included in the three datasets. 

We also decided to perform our experiments in two different settings. 
In the first one, we train all the parameters of every architecture which correspond to standard fine-tuning. 
In the second setting, we decided to freeze the weights of the underlying transformers. In this case, the different transformers are applied once to produce frozen representations. Then, the training will only happen on the top level parameters of the different architectures. This means that we will only train classifiers for all the baselines, and the classifier along with the attention layer of our AoSE architecture. Training our linear attention layer do not take more time than training the classifier itself.

Regarding the other experimental settings, we set the minimum number of tokens per sentence for our AoSE system to 5 and the maximum value to 250. The document representation in the frozen setting is average pooling as it allowed us to obtain better results for all models. When fine-tuning, we use the default pooling strategy implemented by each model. 
For all models and all datasets, we set the learning rate to $2e^-5$ and the batch size to 16. When the memory of our GPUs is exceeded, we reduce the batch size but use gradient accumulation to simulate the same parameters update as with a batch size of 16. 
When freezing the transformers, we train all models for 50 epochs on each dataset. When fine-tuning, we train all models for 20 epochs on IMDB and News Groups, and for 10 epochs on MIND.

\subsection{Evaluations}

Tables \ref{tab:IMDB}, \ref{tab:MIND} and \ref{tab:20NG} present the results obtained by the different models on each dataset\footnote{For more information about the {\em Max seq. length} mentioned in these tables, see subsection \ref{sec:settings}.}.
Overall, we can observe that Longformer, XLNet and AoSE obtain better accuracies on long documents when compared with the remaining baselines. When fine-tuning the transformers, there is no clear best model between them across the three datasets, which joins the conclusions drawn in \cite{2022-efficient}. However, when freezing the transformers, our AoSE model obtains systematically the best results. We believe that this setting is useful for applications that use the same underlying transformer as encoder for multiple tasks or s imply for applications that cannot afford expensive training. Finally, being much smaller than the other models, SMITH obtains the worst accuracies in both settings across all the datasets. 


\begin{table*}
\caption{Results on IMDB.}
\begin{tabular}{lccccccccc}
\hline
& \textbf{Max} & \multicolumn{4}{c}{\textbf{Fine-tuning}} & \multicolumn{4}{c}{\textbf{Freezing}} \\ \cline{3-10}
\textbf{Model} & \textbf{seq.} & \textbf{ Time } & \textbf{ Acc.} & \textbf{ Acc.} & \textbf{ Acc.} & \textbf{ Time } & \textbf{ Acc.} & \textbf{ Acc.} & \textbf{ Acc.} \\
& \textbf{length } & (hh:mm) & all & $<$=512 & $>$512 & (hh:mm) & all & $<=$512 & $>$512 \\
\hline
Roberta & 512 & 05:40 & 95.2 & 95.6 & 93.0 & 00:12 & 91.8 & 92.2 & 88.4 \\
SMITH & 2048 & 04:28 & 91.0 & 91.0 & 90.8 & 00:29 & 74.0 & 74.2 & 73.0 \\
Longformer & 4096 & 22:50 & 95.6 & 95.6 & 95.4 & 00:30 & 92.0 & 92.0 & 91.2 \\
XLNet & 4096 & 22:20 & 95.6 & 95.5 & 95.7 & 00:29 & 92.2 & 92.2 & 91.8 \\
AoSE & 8192 & 15:50 & \textbf{95.7} & \textbf{95.7} & \textbf{95.8} & 00:35 & \textbf{93.2} & \textbf{93.2} & \textbf{93.1} \\ \hline
\end{tabular}
\label{tab:IMDB}
\end{table*}

\begin{table*}
\caption{Results on MIND.}
\begin{tabular}{lccccccccc}
\hline
 & \textbf{Max} & \multicolumn{4}{c}{\textbf{Fine-tuning}} & \multicolumn{4}{c}{\textbf{Freezing}} \\ \cline{3-10}
\textbf{Model} & \textbf{seq.} & \textbf{ Time } & \textbf{ Acc.} & \textbf{ Acc.} & \textbf{ Acc.} & \textbf{ Time } & \textbf{ Acc.} & \textbf{ Acc.} & \textbf{ Acc.} \\
 & \textbf{length } & (hh:mm) & all & $<$=512 & $>$512 & (hh:mm) & all & $<=$512 & $>$512 \\
\hline
Roberta & 512 & 09:20 & 83.1 & 80.7 & 85.8 & 00:45 & 77.4 & 73.8 & 81.6 \\
SMITH & 2048 & 13:30 & 80.8 & 77.4 & 84.8 & 00:52 & 76.0 & 71.6 & 81.1 \\
Longformer & 4096 & 66:45 & \textbf{84.1} & \textbf{80.7} & \textbf{88.0} & 02:05 & 77.7 & 73.7 & 82.3 \\
XLNet & 4096 & 81:58 & 83.4 & 79.8 & 87.3 & 04:12 & 78.3 & 74.2 & 82.9 \\
AoSE & 8192 & 83:10 & 83.5 & 79.8 & 87.5 & 04:58 & \textbf{79.1} & \textbf{75.0} & \textbf{83.9} \\ \hline
\end{tabular}
\label{tab:MIND}
\end{table*}

\begin{table*}
\caption{Results on 20 News Groups.}
\begin{tabular}{lccccccccc}
\hline
 & \textbf{Max} & \multicolumn{4}{c}{\textbf{Fine-tuning}} & \multicolumn{4}{c}{\textbf{Freezing}} \\ \cline{3-10}
\textbf{Model} & \textbf{seq.} & \textbf{ Time } & \textbf{ Acc.} & \textbf{ Acc.} & \textbf{ Acc.} & \textbf{ Time } & \textbf{ Acc.} & \textbf{ Acc.} & \textbf{ Acc.} \\
 & \textbf{length } & (hh:mm) & all & $<$=512 & $>$512 & (hh:mm) & all & $<=$512 & $>$512 \\
\hline
Roberta & 512 & 02:15 & 72.5 & 71.5 & 83.2 & 00:05 & 63.7 & 62.6 & 75.0 \\
SMITH & 2048 & 02:30 & 60.0 & 58.6 & 74.5 & 00:07 & 56.4 & 55.0 & 71.8 \\
Longformer & 4096 & 10:20 & 72.7 & 71.4 & \textbf{84.3} & 00:10 & 64.1 & 62.8 & 77.7 \\
XLNet & 4096 & 13:36 & \textbf{72.8} & \textbf{71.7} & 84.2 & 00:17 & 65.3 & 63.8 & 79.5 \\
AoSE & 8192 & 14:30 & 72.7 & 71.5 & 83.9 & 00:25 & \textbf{66.0} & \textbf{64.4} & \textbf{79.9} \\ \hline
\end{tabular}
\label{tab:20NG}
\end{table*}

Regarding the training speed, we can observe that the frozen setting reduces drastically the training time. 
Since the {\em Max seq. length} differs from one model to the other, comparing their training times is not straightforward. However, we can use the IMDB dataset for a fair comparison between Longformer, XLNet and AoSE as all IMDB documents can be modeled entirely without any truncation by these three models\footnote{the longest IMDB document has less than 4096 tokens.}. In this case, AoSE is faster than the two other models in the fine-tuning setting, but slightly slower in the frozen setting. 

\subsection{Ablation Study}

We conduct an ablation study to further investigate the relevancy of the studied architecture over simpler baselines. We compare the results obtained by the selected sentence transformer alone (S-Roberta) with different versions of our AoSE architecture that use the same sentence transformer as their first component. The new AoSE versions are fed with sequences that are truncated after 512, 1024, 2048 and 4096 tokens respectively.
Table \ref{tab:ablation} shows the obtained results on the IMDB dataset. 
It appears that the AoSE architecture is relevant and benefits from increasing the sequence length especially on long documents (that have more than 512 tokens). We also observe a slight improvement on short documents (having less than 512 tokens) in the frozen setting, which may be explained by a better attention layer after training on the additional sentences that appear at the end of long documents. 

\begin{table*}
\caption{Ablation study conducted on IMDB: (i) S-Roberta refers to the chosen sentence transformer used alone and applied to the whole sequence (truncated after 512 tokens); (ii) AoSE-xxx refers to the application of the proposed architecture that also uses S-Roberta (input sequences are truncated after xxx tokens).}
\begin{tabular}{lccccccc}
\hline
\multirow{3}{*}{\textbf{Model}}  & \textbf{ Max } & \multicolumn{3}{c}{\textbf{ Fine-tuning}} & \multicolumn{3}{c}{\textbf{Freezing}} \\ \cline{3-8}
 & \textbf{ sequence } & \textbf{ Acc. } & \textbf{ Acc. } & \textbf{ Acc. } & \textbf{ Acc. } & \textbf{ Acc. } & \textbf{ Acc. } \\
 & \textbf{length} & all & $<$=512 & $>$512 & all & $<$=512 & $>$512 \\
\hline
S-Roberta & 512 & 95.3 & 95.6 & 92.8 & 92.2 & 92.7 & 89.1 \\
AoSE-512 & 512 & 95.4 & 95.7 & 93.0 & 92.6 & 93.0 & 89.9 \\
AoSE-1024 & 1024 & 95.7 & 95.7 & 95.7 & 93.0 & 93.1 & 92.5 \\
AoSE-2048 & 2048 & 95.7 & 95.7 & 95.8 & 93.1 & 93.1 & 92.9 \\
AoSE-4096 & 4096 & 95.7 & 95.7 & 95.8 & 93.2 & 93.2 & 93.1 \\
\hline
\end{tabular}
\label{tab:ablation}
\end{table*}

\subsection{Impact of the chosen sentence transformers}

Finally, we evaluate the impact of the chosen sentence transformer on the final results. Table \ref{tab:ST} below shows the results obtained with three different sentence transformers in the same two settings presented earlier (fine-tuning and frozen). In addition to the already evaluated S-Roberta, two other sentence transformers have been included here: S-BERT\footnote{\url{https://huggingface.co/sentence-transformers/bert-base-nli-mean-tokens}} and S-MPNet\footnote{\url{https://huggingface.co/sentence-transformers/paraphrase-mpnet-base-v2}}. Again, the IMDB dataset is used to conduct these experiments.

\begin{table*}
\caption{Results of different sentence transformers used either alone and inside our architecture on the IMDB dataset. Ao(S-Roberta) refers to attention over S-Roberta, the same model used in our previous experiments. Ao(S-BERT) refers to attention over S-BERT. Ao(S-MPNet) refers to attention over S-MPNet.}
\begin{tabular}{lccccccc}
\hline
\multirow{3}{*}{\textbf{Model}}  & \textbf{ Max } & \multicolumn{3}{c}{\textbf{ Fine-tuning}} & \multicolumn{3}{c}{\textbf{Freezing}} \\ \cline{3-8}
 & \textbf{ sequence } & \textbf{ Acc. } & \textbf{ Acc. } & \textbf{ Acc. } & \textbf{ Acc. } & \textbf{ Acc. } & \textbf{ Acc. } \\
 & \textbf{length} & all & $<$=512 & $>$512 & all & $<$=512 & $>$512 \\
\hline
S-Roberta & 512 & 95.3 & 95.6 & 92.8 & 92.2 & 92.7 & 89.1 \\
Ao(S-Roberta) & 8192 & \textbf{95.7} & \textbf{95.7} & \textbf{95.8} & \textbf{93.2} & \textbf{93.2} & \textbf{93.1} \\ \hdashline
S-BERT & 512 & 94.0 & 94.5 & 86.2 & 87.0 & 87.9 & 82.1 \\
Ao(S-BERT) & 8192 & \textbf{94.8} & \textbf{94.8} & \textbf{94.6} & \textbf{90.7} & \textbf{90.7} & \textbf{90.4} \\ \hdashline
S-MPNet & 512 & 95.2 & 95.5 & 93.3 & 91.7 & 92.2 & 88.3 \\
Ao(S-MPNet) & 8192 & \textbf{95.6} & \textbf{95.5} & \textbf{96.0} & \textbf{93.3} & \textbf{93.3} & \textbf{92.8} \\ \hline
\end{tabular}
\label{tab:ST}
\end{table*}

Each model is first applied directly to the whole sequence (truncated after 512 tokens)\footnote{Initial evaluations have shown that applying sentence transformers directly to the whole document gives better results than applying them to each sentence and then averaging the different embeddings.}. Then, it is used inside the studied architecture that is able to handle up to 8192 tokens. 
Overall, it appears that S-BERT obtains lower results than S-Roberta and S-MPNet either when used alone or inside our architecture. Therefore, we can say that the choice of the sentence transformer has an important impact on the final results. 
But more importantly, we observe that using the same models inside our AoSE architecture allow to improve the results of all models. This improvement is observable in both settings (fine-tuning and freezing), regardless of the underlying sentence transformer. 


\section{Conclusion}

In this paper, we investigated the relevancy of pre-trained sentence transformers for long document classification. To our knowledge this is the first time these pre-trained models are used to handle long documents. 
To do so, we combine the sentence representations using a trainable attention layer to give high weights to important sentences. We have shown that this simple method is competitive when compared with current state-of-the-art models in the standard fine-tuning mode. We also considered another mode where the underlying transformers are frozen, which allows to speed-up the training and to share the same underlying transformer between different applications. In this case, the AoSE architecture obtains better results. Additional experiments have shown an improvement over the direct application of the same sentence transformers. 
Finally, relying on pre-trained sentence transformers allows to easily extend our architecture to different domains and languages. For example, we can simply replace the English sentence transformer used in this paper with a multilingual one\footnote{For example: \url{https://huggingface.co/sentence-transformers/stsb-xlm-r-multilingual}} to handle multilingual texts, whereas XLNet and Longformer need to be pre-trained again on multilingual datasets.

\bibliography{custom}




\end{document}